\documentclass[letterpaper, 10pt, conference]{ieeeconf}
\IEEEoverridecommandlockouts
\newcommand{\ShowNotes}{}
\usepackage[utf8]{inputenc}
\usepackage[english]{babel}

\usepackage{comment}
\usepackage{xspace}
\usepackage{graphicx}
\usepackage{overpic}
\usepackage{svg}
\usepackage{xcolor}
\usepackage{tikz}
\usetikzlibrary{arrows.meta, positioning, shapes.geometric, fit, backgrounds, decorations.pathreplacing, calc}
\usepackage{pgfplots}
\pgfplotsset{compat=1.18}
\usepackage{bm}
\usepackage[hang,flushmargin]{footmisc}
\usepackage{adjustbox}
\usepackage{etoolbox}
\usepackage{environ}
\usepackage{pbalance}
\usepackage{flushend}
\usepackage{mathtools}
\usepackage[normalem]{ulem}

\usepackage{pifont}
\usepackage{tabularx}
\usepackage{multirow}
\usepackage{booktabs}
\usepackage{colortbl}
\usepackage{float}
\usepackage{placeins}
\usepackage{array}
\usepackage{threeparttable}

\usepackage{amsmath}
\usepackage{amssymb}
\usepackage{amsthm}
\usepackage{siunitx}

\makeatletter
\patchcmd{\@makecaption}{\scshape}{}{}{}
\newcommand{\onetagright}{\tagsleft@false}
\makeatother

\usepackage{subcaption}
\newtheoremstyle{main}
{1em}                                                %
{1em}                                                %
{\itshape}                                           %
{0pt}                                                %
{\scshape}                                           %
{\\*}                                                %
{2pt}                                                %
{\thmname{#1}\thmnumber{ #2}: \thmnote{\itshape #3}} %

\usepackage[linesnumbered,ruled,noend]{algorithm2e}
\newcommand{\removelatexerror}{\let\@latex@error\@gobble}

\let\labelindent\relax
\usepackage[inline]{enumitem}

\usepackage[
	activate   = {true},
	protrusion = false,
	expansion  = true,
	kerning    = true,
	spacing    = true,
	tracking   = false,
	auto       = true,
	selected   = true,
	factor     = 1000,
	stretch    = 10,
	shrink     = 10,
]{microtype}

\usepackage{csquotes}
\usepackage[
	maxbibnames=7,
	minbibnames=7,
	maxcitenames=2,
	natbib=true,
	bibstyle=ieee,
	citestyle=numeric-comp,
	backend=biber,
	sorting=none,
	giveninits=true,
	url=false,
	doi=false,
	eprint=false,
	isbn=false,
]{biblatex}

\definecolor{purduegold}{HTML}{C28E0E} %

\makeatletter
\let\NAT@parse\undefined
\makeatother
\usepackage[pdfa,colorlinks,bookmarksopen,bookmarksnumbered,allcolors=purduegold]{hyperref}
\usepackage{bookmark}

\usepackage[nameinlink,capitalise]{cleveref}
\crefname{line}{line}{lines}
\crefname{figure}{Fig.}{Figs.}
\Crefname{figure}{Fig.}{Figs.}
\crefname{equation}{Eq.}{Eqs.}
\Crefname{equation}{Eq.}{Eqs.}
\crefname{section}{Sec.}{Secs.}
\Crefname{section}{Sec.}{Secs.}
\crefname{definition}{Def.}{Defs.}
\Crefname{definition}{Def.}{Defs.}
\crefname{algorithm}{Alg.}{Algs.}
\Crefname{algorithm}{Alg.}{Algs.}
\crefname{assumption}{Asm.}{Asms.}
\Crefname{assumption}{Asm.}{Asms.}
\crefname{subassumption}{Asm.}{Asms.}
\Crefname{subassumption}{Asm.}{Asms.}
\crefname{problem}{Problem}{Problems}
\Crefname{problem}{Problem}{Problems}

\makeatletter
\newcommand\footnoteref[1]{\protected@xdef\@thefnmark{\ref{#1}}\@footnotemark}
\makeatother

\graphicspath{ {./figs/} }

\usepackage{soul}
\usepackage{placeins}
\definecolor{darkred}{rgb}{0.7,0.1,0.1}
\definecolor{darkgreen}{rgb}{0.1,0.7,0.1}
\definecolor{cyan}{rgb}{0.7,0.0,0.7}
\definecolor{dblue}{rgb}{0.2,0.2,0.8}
\definecolor{maroon}{rgb}{0.76,.13,.28}
\definecolor{burntorange}{rgb}{0.81,.33,0}
\definecolor{tealblue}{rgb}{0.212,0.459, 0.533}

\definecolor{mypink}{rgb}{0.93359375, 0.62109375, 0.83984375}

\definecolor{pp}{rgb}{0.43921569, 0.18823529, 0.62745098}
\definecolor{rr}{rgb}{0.5254902 , 0.00784314, 0.12941176}
\definecolor{bb}{rgb}{0.09019608, 0.23529412, 0.37647059}
\definecolor{yy}{rgb}{0.49803922, 0.3372549 , 0.0}
\definecolor{gg}{rgb}{0.02352941, 0.3372549 , 0.17647059}

\ifdefined\ShowNotes
  \newcommand{\colornote}[3]{{\color{#1}\bf{#2: #3}\normalfont}}
\else
  \newcommand{\colornote}[3]{}
\fi

\usepackage{amsmath,amsfonts,bm}
\usepackage{amsthm,amssymb}

\def\1{\bm{1}}

\newcommand{\beas}{\begin{eqnarray*}}
\newcommand{\eeas}{\end{eqnarray*}}
\newcommand{\bea}{\begin{eqnarray}}
\newcommand{\eea}{\end{eqnarray}}
\newcommand{\bes}{\begin{equation*}}
\newcommand{\ees}{\end{equation*}}
\newcommand{\be}{\begin{equation}}
\newcommand{\ee}{\end{equation}}

\makeatletter
\usepackage{xspace}
\def\@onedot{\ifx\@let@token.\else.\null\fi\xspace}
\DeclareRobustCommand\onedot{\futurelet\@let@token\@onedot}

\newcommand{\figref}[1]{Fig\onedot~\ref{#1}}
\newcommand{\equref}[1]{Eq\onedot~\eqref{#1}}

\newcommand{\tabref}[1]{Tab\onedot~\ref{#1}}

\DeclareMathAlphabet{\mathsfit}{\encodingdefault}{\sfdefault}{m}{sl}
\SetMathAlphabet{\mathsfit}{bold}{\encodingdefault}{\sfdefault}{bx}{n}

\newcommand{\R}{\mathbb{R}}

\title{%
\LARGE \bf SkipVLA: Skipping VLA Steps with Classical Planning\\ for Fast Robot Manipulation
}

\newif\ifanonymous
\anonymousfalse

\ifanonymous
  \author{Anonymous Author(s)}
\else
    \author{
    Kaivalya Agrawal, Md Ashiqur Rahman, Raymond A. Yeh, and Zachary Kingston%
    \thanks{The authors are with the Department of Computer Science, Purdue University, {\tt \{agraw201, rahman79, rayyeh, zkingston\}@purdue.edu}.
    }}
\fi

\begin{document}
\maketitle
\thispagestyle{empty}
\pagestyle{empty}

\begin{abstract}
Vision-Language-Action (VLA) models are a class of generalist robot policies that map camera images and language instructions directly to robot actions.
While promising, these models remain slow at test time, particularly for long-horizon tasks that require many queries to the policy. Recent efforts reduce VLA latency by distilling smaller models, overlapping asynchronous action chunks, or pairing the VLA with a fast low-level policy, but still run a learned policy for the entire task. In contrast to VLA, classical motion planners quickly find collision-free motions, but require an explicit goal and have no semantic understanding of the task.
In this work, we present SkipVLA, a hybrid policy that combines a pretrained VLA with a classical motion planner, using the planner for free-space motion and querying the VLA only for contact-rich skills such as grasping and placing.
SkipVLA reuses the frozen vision-language backbone of the VLA to predict a target pose for each planned motion, and learns this predictor without additional demonstrations introduced into the system by using what was already learnt by the large VLA.
We evaluate SkipVLA with three VLAs on 13 LIBERO tasks in simulation and three pick-and-place tasks on a physical 6-DoF YAM arm, demonstrating up to 2.5x faster task completion and significantly lower energy consumption while achieving the same task success rate.

\end{abstract}

\section{Introduction}
\label{sec:intro}
Vision-Language-Action (VLA) models~\cite{rt22023, pi05} have emerged as a promising approach for generalist robot manipulation. By pairing pretrained vision-language backbones with expressive action heads, VLAs map raw camera observations and natural-language task instructions directly to robot actions. This unified formulation enables open-vocabulary instruction following, cross-scene generalization, and long-horizon task execution without handcrafted heuristics.

Despite their expressive capability, VLAs face a severe practical limitation: high inference latency. Generating each action chunk requires forward passes through multi-billion-parameter networks, resulting in latencies that far exceed the robot's control cycle. In long-horizon tasks requiring dozens or hundreds of queries, this overhead leads to sluggish execution, high energy consumption on edge compute platforms, and compromised reactive safety.

\begin{figure}[t]
    \centering
    \includegraphics[width=\linewidth]{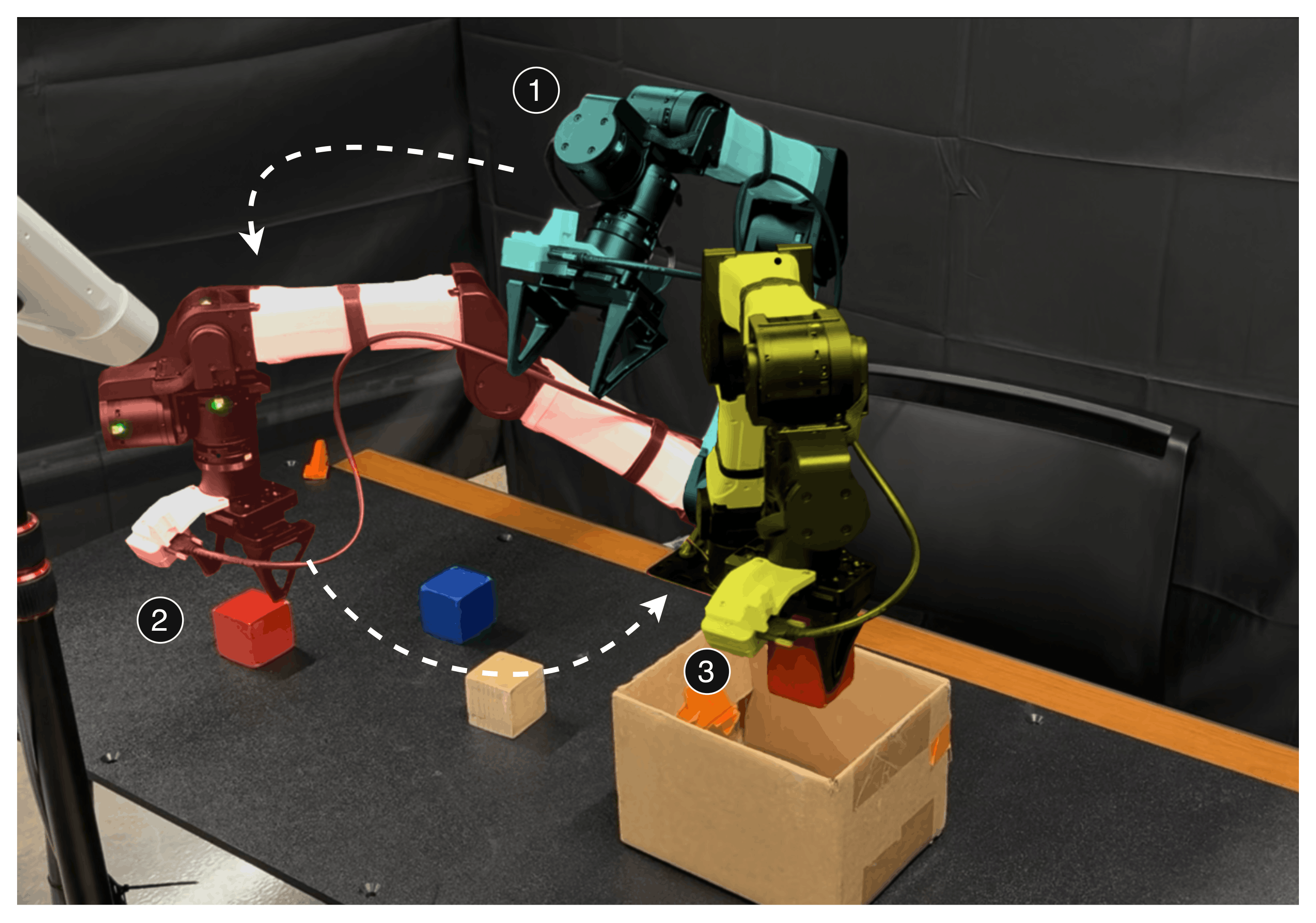}
    \caption{Our hybrid method on a real-world pick-and-place task, shown at three points in time. White arrows are free-space transit executed by the planner module. \textbf{(\ding{182}, cyan)} From the start pose, the planning module predicts a hover target above the red block and moves the arm to it. \textbf{(\ding{183}, red)} The VLA expert module performs the grasp. When the gripper closes, control returns to the planner, which moves the arm to a predicted hover target above the box. \textbf{(\ding{184}, yellow)} The VLA expert places the block in the box.}
    \vspace{-0.5cm}
    \label{fig:my_single_column_image}
\end{figure}

Prior work to reduce VLA latency typically distills smaller models~\cite{wen_tinyvla_2025}, overlaps asynchronous action chunks~\cite{black_real-time_2025}, or pairs the VLA with a fast low-level policy in a hierarchical setup~\cite{han_dual_2024}. However, each introduces notable trade-offs: distilled models frequently suffer performance degradation; asynchronous chunking risks executing stale actions that push the policy out of distribution; and hierarchical setups require training a completely new policy network from scratch for specialized subtasks. Crucially, all of these methods treat manipulation as a monolithic end-to-end learning problem, querying large neural networks even during simple, unconstrained motions.

In reality, manipulation tasks exhibit a natural physical decomposition: they alternate between short, \emph{contact-rich manipulation} skills (e.g., grasping, reorienting, or placing) and long stretches of \emph{free-space transit} (e.g., reaching toward an object or navigating to a receptacle). Contact-rich manipulation requires adaptive, closed-loop visuomotor feedback to manage complex physical contacts, a domain where learned VLAs excel. In contrast, free-space transit is a purely geometric and kinematic problem that requires collision avoidance but minimal semantic reasoning.

Classical motion planners, such as sampling-based algorithms~\cite{vamp_2024} and trajectory optimizers~\cite{curobo_v2}, solve this geometric problem rapidly, generating collision-free trajectories in microseconds to milliseconds with formal safety guarantees. However, classical planners are semantically blind, i.e., they require explicit goal poses and cannot infer them from natural language. Methods that prompt foundation models for zero-shot spatial keypoints~\cite{huang2023voxposer, huang2024rekep, liu2024moka} and only rely on geometric optimization for the entire task struggle with the delicate contact dynamics that VLAs naturally handle.

This motivates a hybrid architecture that reserves the VLA for contact-rich interactions, offloads free-space transit to a fast classical planner, and uses the VLA's frozen representations to predict the planner's target poses.

To this end, we introduce SkipVLA, a modular hybrid policy that dynamically interleaves a pretrained VLA with a classical motion planner. SkipVLA scores candidate 3D object bounding boxes, conditioned on the base VLA's frozen vision-language features, to determine collision-free hover targets. A classical planner navigates the robot through free space to the target pose, then hands control to the VLA for contact-rich manipulation. Once the manipulation completes as signaled by a gripper state transition, control returns to the planner. We train the target predictor using self-supervision extracted from existing demonstration trajectories at gripper events, requiring no manual annotations, architectural changes, or policy retraining.
{\bf \noindent Our main contributions are:} %
\begin{itemize}
    \item  We present SkipVLA, a plug-and-play framework that accelerates manipulation by routing free-space transit to classical motion planners and reserving pretrained VLAs solely for contact-rich skills.
    \item  We develop a lightweight scoring head that uses the VLA's frozen vision-language tokens to predict 3D target poses and can be trained on an existing imitation learning dataset with no additional annotation.
    \item  We evaluate SkipVLA across 13 tasks on the LIBERO benchmark in simulation and three physical tasks on a 6-DoF YAM arm, spanning three distinct VLA backbones ($\pi_{0.5}$, SmolVLA, and MolmoAct2) and multiple motion planners (VAMP and cuRoboV2).
    \item  SkipVLA achieves up to a $2.5\times$ wall-clock speedup ($59.6\%$ latency reduction) and cuts compute energy by over $52\%$ on an NVIDIA Jetson Thor. 
\end{itemize}

\section{Related Work}
\label{sec:related_work}
\begin{figure*}[htbp]
\centering
\includegraphics[width=\textwidth]{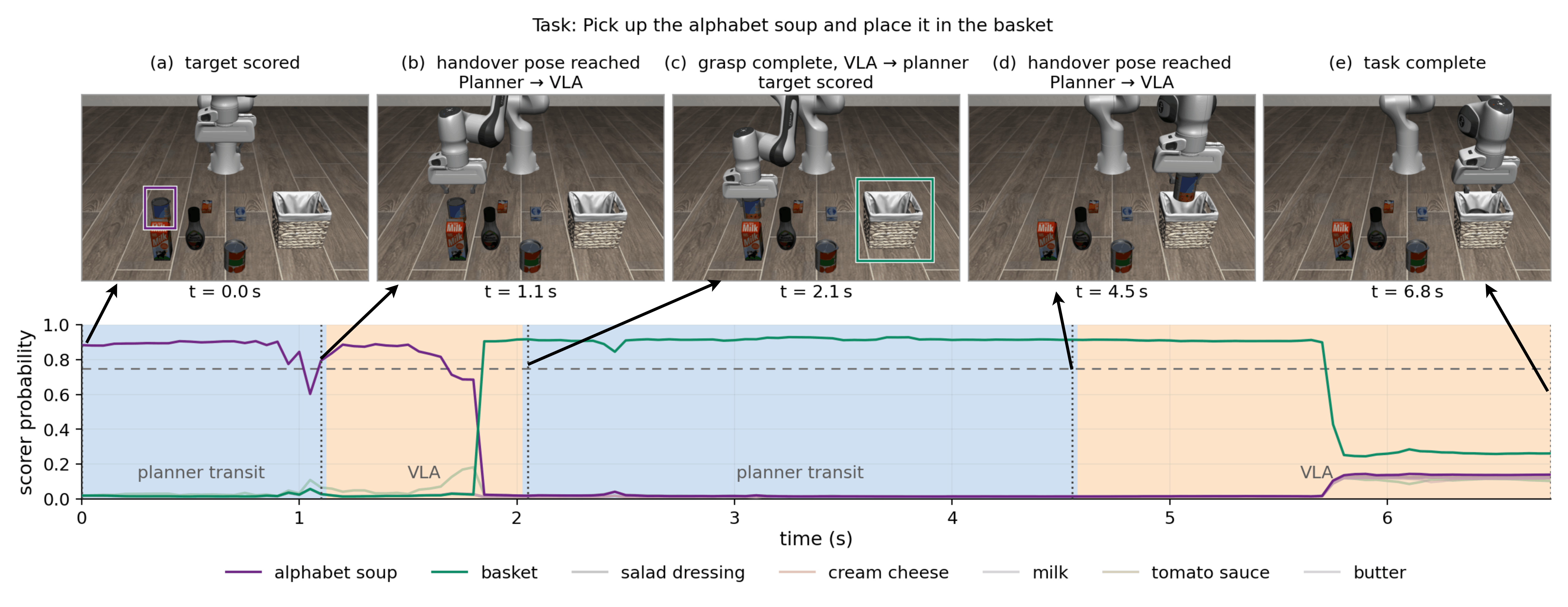}
\caption{\textbf{SkipVLA on a single LIBERO-Object episode.} Top: five keyframes from one successful rollout of $\pi_{0.5}$ + SkipVLA with the VAMP planner. \textbf{(a)} The scoring module selects a target object and pose. \textbf{(b)} VAMP's plan is followed from start to target pose and control is handed over to VLA. \textbf{(c)} The VLA completes the skill of grasping and the control is handed over to SkipVLA which scores the next target. \textbf{(d)} VAMP's plan is followed from start to target pose and control is handed over to VLA. \textbf{(e)} The VLA completes the skill of placing and task is marked complete. Bottom: the scoring module's probability over all seven tracked object bounding boxes for the same episode, with background shading showing which policy holds control at each step: planner transit (blue) and VLA (orange). Control returns to the planner at exactly the grasp and release transitions. Over the 6.8s episode, the planner drives 54\% of the task. The deployed system queried the scorer twice; the continuous probability curves are an offline re-scoring of every frame with a fresh latent to show the change in the scoring signal throughout the trajectory.}
\label{fig:pipeline}
\end{figure*}

{\bf Vision-language-action models.} Following the success of Vision-Language Models (VLMs)~\cite{bai2025qwen3,team2026gemma} in scene understanding and cross-modal semantic alignment, substantial research has aimed to construct open-ended robotic policies that unify perception and control within a single end-to-end framework, known as VLA models. Recently, VLAs have expanded to incorporate generative architectures for better generalization and performance, such as diffusion- and flow-matching-based action heads~\cite{pi0, pi05, octo2024}, to capture complex, multimodal action distributions. While these models have been successful in executing semantically aligned, long-horizon tasks, their practical deployment is severely bottlenecked due to their massive parameter counts, making them unsuitable for high-frequency real-world control.

{\bf Efficient VLA architectures.} Extensive research has been conducted to reduce the high inference latency of VLAs. Large VLAs have been distilled into smaller variants to reduce parameter counts and improve inference speed~\cite{wen_tinyvla_2025, prasad_consistency_2024}.
However, such models often exhibit drops in performance due to distribution shift and still remain relatively slow for high-frequency control. Other works employ action chunking~\cite{black_real-time_2025, black_training-time_2025, act2023} to overlap asynchronously inferred action chunks, effectively hiding execution pauses due to slow inference. These approaches introduce staleness in the inferred chunk, and interleaving action chunks can push the policy off distribution. Finally, hierarchical methods introduce a low-latency, smaller policy module that operates asynchronously alongside a large VLA~\cite{han_dual_2024, chen_fast--slow_2025, zhang_hirt_2025, bu_towards_2024, chen2026streamvla}, reusing the visual representations of the larger network. 

{\bf Classical motion planning.} Classical planning approaches like Sampling-Based Motion Planning algorithms (SBMP) and trajectory optimization methods have long proven to be successful in solving complex navigation and manipulation tasks. Modern implementations of SBMPs are highly energy- and time-efficient, capable of generating collision-free trajectories in microseconds~\cite{vamp_2024, wilson_ral25}. Similarly, gradient-based trajectory optimization frameworks~\cite{Ratliff-2009, trajopt2014} and GPU-accelerated solvers~\cite{curobo_v2} calculate dynamically feasible motion in real time. Despite their safety guarantees and low-latency planning, classical planners are semantically blind. They operate purely on coordinate geometry and require well-defined goal states or objective functions.

{\bf Combining learning with classical planning.} 
Recent works have attempted to combine the semantic reasoning of foundation models with the reliability of classical planners. One line of research, including VoxPoser~\cite{huang2023voxposer} and ReKep~\cite{huang2024rekep}, prompts VLMs to output spatial cost maps or relational constraints that are then solved by trajectory optimization. However, because these methods rely entirely on zero-shot geometric optimization rather than learned action priors, they struggle to execute dynamic, contact-rich manipulation that cannot be represented as a cost function. 

A parallel line of work attempts to integrate classical planning directly with trained behavioral policies like LiLo-VLA~\cite{lilo2026} and MPVI~\cite{choe2026makevlarobustdata}. However, these methods suffer from distinct architectural bottlenecks. Frameworks like MPVI treat the motion planner primarily as a recovery mechanism, using it to veer the policy back to in-distribution only after it drifts out-of-distribution. Other works, like LiLo-VLA, rely on pre-defined action sequences before performing simple skill chaining.

In contrast to these frameworks, SkipVLA interleaves a generalist VLA with classical sampling-based motion planning at planning and execution time. By leveraging the VLA's pretrained representations strictly for target pose prediction, we dynamically hand off all free-space navigation in between atomic skills to the fast classical planner. This achieves massive latency reductions and guarantees collision-free transit without compromising on task success rate or introducing shifts in policy and data distribution within the VLA and builds on top of the base VLA architecture.

\section{Background}
\label{sec:background}
In this section, we introduce the background and notation used in this work.

{\bf VLA.} We denote a generalist VLA robot policy as $\pi^{\text{vla}}$. This policy maps an observation $\mathcal{O}_s$, a natural-language instruction $L$ (task description), and an optional proprioceptive state $\mathbf{p}_s$ to a robot action 
\begin{equation}
\label{eqn:vla_policy}
    a_s = \pi^{\text{vla}}(\mathcal{O}_s, L, \mathbf{p}_s)
\end{equation}
at step $s$. 
The robot's proprioceptive state $\mathbf{p}_s$ consists of the joint angles and gripper state (with corresponding end-effector pose $\mathbf{p}_s^{ee}$). The observation $\mathcal{O}_s$ includes an image and depth map from a workspace viewpoint RGBD camera and an image from a wrist-mounted RGB camera. 

{\bf Trajectory.} We use the term \emph{trajectory} to denote a sequence of proprioceptive states $ \tau$ as
\begin{equation}
\label{eqn:trajectory}
    \tau = (\mathbf{p}_1, \mathbf{p}_2, \dots, \mathbf{p}_{N}),
\end{equation}
 where $N = |\tau|$ is the trajectory length and $a_s=\mathbf{p}_{s+1} - \mathbf{p}_s$ is the action taken at step $s$ that changes the robot's proprioceptive state from $\mathbf{p}_s$ to $\mathbf{p}_{s+1}$. 

{\bf Action chunks.} Different VLA architectures generate actions at different granularities. A VLA when queried at step $s$ can produce an action \emph{chunk} of $K$ future actions
\begin{equation}
\label{eqn:action_chunk}
    A_s = (a_s, a_{s+1}, \dots, a_{s+K-1}) = \pi^{\text{vla}}(\mathcal{O}_s, L, \mathbf{p}_s).
\end{equation}
For a given VLA, the action chunk horizon $K \ge 1$ is a fixed constant. Therefore, the VLA is queried every $K$ steps. We denote the set of these \emph{query steps} by $\mathcal{Q} = \{1, 1+K, 1+2K, \dots\} \subset \{1, \dots, N\}$, with $|\mathcal{Q}| = \lceil N/K \rceil$.

{\bf Execution time.} The total execution time for a trajectory $\tau$ of length $N = |\tau|$ is governed by two factors: the control period $\Delta t_{\text{ctrl}} = \frac{1}{f_{\text{ctrl}}}$, dictated by the sampling frequency $f_{\text{ctrl}}$ of the robot controller, and the inference latency $\Delta t_{\text{lat}}$, which is the processing time required by the VLA to generate each action chunk.

The total execution time of the trajectory $\tau$ generated by policy $\pi$ is given by:
\begin{equation}
\label{eqn:exec_time}
    \mathcal{T}(\tau) = N \cdot \Delta t_{\text{ctrl}} + \sum_{s \in \mathcal{Q}} \Delta t_{\text{lat}} = N \Bigl( \Delta t_{\text{ctrl}} + \frac{\Delta t_{\text{lat}}}{K} \Bigr),
\end{equation}
where $\Delta t_{\text{ctrl}}$ and $\Delta t_{\text{lat}}$ are the control period and inference latency of $\pi$, and $|\mathcal{Q}| = N/K$ is the number of query steps (assuming $N$ is a multiple of $K$ for simplicity). For a generalist VLA with a large number of parameters, $\Delta t_{\text{lat}}$ is often the bottleneck of the execution time ($\Delta t_{\text{lat}} \gg \Delta t_{\text{ctrl}}$).

{\bf Imitation learning.} In robot manipulation, policies are mainly trained via imitation learning on a static dataset $\mathcal{D} = \{(\tau^{e,i},\{\mathcal{O}\}^i, L^i)\}$ of expert demonstrations.
The objective trains the policy to imitate expert actions step-by-step:
\begin{equation}
\label{eqn:bc_step}
    \min_{\pi} \; \mathbb{E}_{(\tau^e,\{\mathcal{O}\}, L) \sim \mathcal{D}} \;\sum_{s=1}^{|\tau^e|} \ell\left(a_s^e,\; \pi(\mathcal{O}_s, L, \mathbf{p}_s)\right),
\end{equation}
where $a_s^e$ is the expert action at step $s$ and $\ell$ is an appropriate loss for the task.

\section{Method}
\label{sec:method}

We propose SkipVLA which augments a generalist VLA policy $\pi^{\text{vla}}$ with a lightweight, low-latency classical control module $\pi^{\text{fast}}$ ($\Delta t_{\text{lat}}^{\text{fast}} \ll \Delta t_{\text{lat}}^{\text{vla}}$). It dynamically routes between classical motion planning (for free-space transit) and a VLA (for contact-rich manipulation) (see~\figref{fig:pipeline}).

\subsection{Problem Formulation and Objective}
\label{sec:method_formulation}
We achieve this hybrid execution through an indicator module $I$ that determines which policy to execute at each timestep. Given an observation $\mathcal{O}_s$, task instruction $L$, and proprioceptive state $\mathbf{p}_s$, the indicator outputs a binary decision $I_s \in \{0, 1\}$, yielding the hybrid policy:
\begin{equation}
\label{eqn:hybrid_policy}
\pi^{\text{hyb}}(\mathcal{O}_s, L, \mathbf{p}_s) \triangleq 
\begin{cases}
\pi^{\text{vla}}(\mathcal{O}_s, L, \mathbf{p}_s),  &  I_s = 0,\\
\pi^{\text{fast}}(\mathcal{O}_s, L, \mathbf{p}_s), & \text{otherwise}.
\end{cases}
\end{equation}
The latency of the hybrid policy at step $s$ is expressed as:
\begin{equation}
\label{eqn:latency}
    \Delta t_s^{\text{lat}} =
    \begin{cases}
        \Delta t_{\text{lat}}^{\text{vla}},  & I_s = 0, \\
        \Delta t_{\text{lat}}^{\text{fast}}, & \text{otherwise.}
    \end{cases}
\end{equation}

{\bf Trajectory-level objective.} To keep the latency low, the $\pi^{\text{fast}}$ module is designed to be lightweight using classical motion planning. However, classical motion planners (such as RRT or PRM) generate collision-free paths that optimize geometric or kinematic criteria rather than mimicking human demonstration trajectories step by step. Consequently, a per-action imitation loss as in \equref{eqn:bc_step} would incorrectly penalize such valid transit trajectories, even when they reach the destination or the target pose safely and faster.

To address this, we use a trajectory-level \emph{imitation learning} objective where, rather than matching every action, the trajectory rolled out by $\pi^{\text{hyb}}$ is constrained to pass through key \emph{waypoints} of the expert trajectory, which ensures successful task completion. To minimize total execution time, the overall objective can be expressed as follows:
\begin{equation}
\label{eqn:objective}
    \min_{\pi^{\text{fast}},\, I} \; \mathbb{E}_{\tau^e \sim \mathcal{D}} \Bigl[ \mathcal{T}(\tau^{\text{hyb}}) + \lambda \cdot \ell_{\text{traj}}(\tau^{\text{hyb}}, \tau^{e}) \Bigr],
\end{equation}
where $\tau^{\text{hyb}}$ is the trajectory rolled out by $\pi^{\text{hyb}}$, $\tau^e$ is the corresponding expert trajectory, $\mathcal{T}(\tau^{\text{hyb}})$ is the execution time, and $\lambda > 0$ is a hyper-parameter of choice.
$\ell_{\text{traj}}(\tau^{\text{hyb}}, \tau^{e}) = 0$ implies the hybrid trajectory passes through all required key waypoints, ensuring task completion.

\begin{figure*}[t]
\centering
\includegraphics[width=\textwidth]{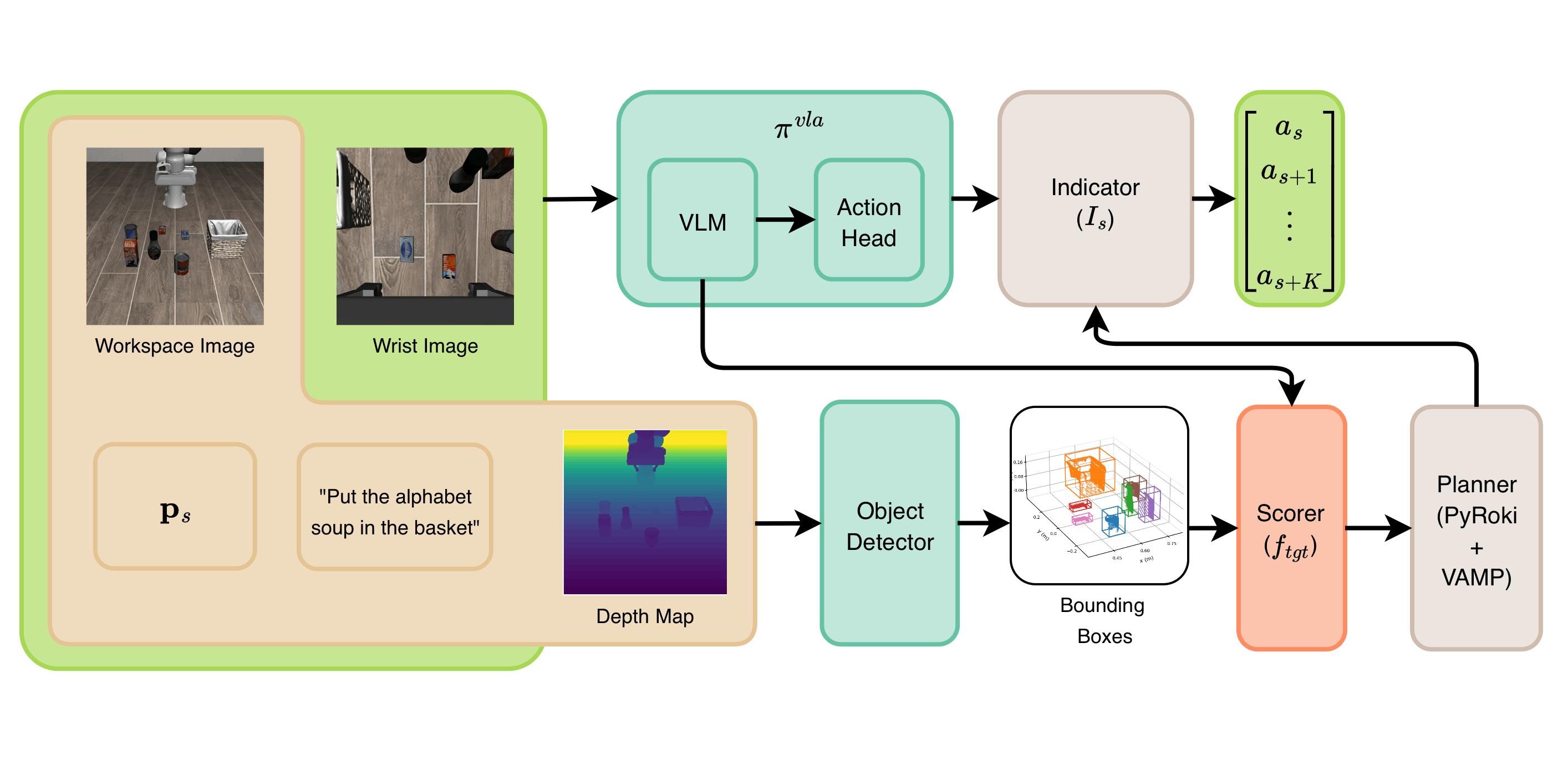}
\vspace{-1.2cm}
\caption{\textbf{Inference pipeline of our proposed SkipVLA.} We represent our input data as $\mathcal{O}_s, L, \mathbf{p}_s$, where $\mathcal{O}_s$ contains a workspace image, a wrist image, and a workspace depth map, $L$ represents the language instruction of the task, and $\mathbf{p}_s$ represents the proprioceptive state. The base expert component $\pi^{vla}$ comprises a VLM and an action head that process the input stream to output an action vector and a vision-language context latent. The object detection module takes in a different subset of inputs and uses FastSAM~\cite{zhao2023fastsam} and DBSCAN to generate object bounding boxes. The scoring module $f_{tgt}$ selects the target object given the bounding boxes and latent context from the VLM. We use PyRoki~\cite{kim2025pyroki} and VAMP to generate an action sequence. Finally, the indicator module $I$ decides which action sequence to execute.
}
\label{fig:infer_pipeline}
\vspace{-.3cm}
\end{figure*}

{\bf Key-waypoints extraction.} The trajectory $\tau$ for completing a task can be partitioned into segments, each associated with a subtask.
Broadly, these subtasks fall into two categories: \emph{active object manipulation}, where the robot's end-effector or gripper performs grasping, reorientation, or other contact-rich interactions with the object; and \emph{reach-and-place} (approach/placement), where the robot primarily moves its end-effector with or without a grasped object.

In more detail, given a trajectory $\tau$, we partition it as $\tau = [\tau^1,  \tau^2 \cdots \tau^n$], where we concatenate each segment $\tau^j$ corresponding to either active object manipulation or reach-and-place.

For an expert trajectory segment $\tau^{e,j}$ corresponding to active object manipulation, the matching segment $\tau^{\text{hyb},j}$ rolled out by $\pi^{\text{hyb}}$ should closely follow the expert at each step to ensure successful completion of the subgoal.
On the other hand, for an expert reach-and-place segment $\tau^{e,j}$, successful completion requires the corresponding segment $\tau^{\text{hyb},j}$ to match only the initial and final end-effector poses. Based on this structure, we define the segment-level loss as follows:
\begin{equation}
\label{eqn:segment_loss}
\begin{aligned}
&\ell_{\text{traj}}(\tau^{\text{hyb},j}, \tau^{e,j}) = \\
&\begin{cases}
    \sum_{s=1}^{N_j^e} d(\tau^{\text{hyb},j}_s, \tau^{e,j}_s)^2,
    & \text{if manipulation,} \\[3pt]
    d(\tau^{\text{hyb},j}_1, \tau^{e,j}_1)^2 + d(\tau^{\text{hyb},j}_{N_j}, \tau^{e,j}_{N_j^e})^2,
    & \text{if reach-and-place,}
\end{cases}
\end{aligned}
\end{equation}
where $\tau^{\text{hyb},j}_s$ and $\tau^{e,j}_s$ represent the end-effector poses at step $s$ for the respective segments. Additionally, $N_j = |\tau^{\text{hyb},j}|$ and $N_j^e = |\tau^{e,j}|$ indicate the lengths of the respective segments, while $d(\cdot,\cdot)$ denotes the chosen distance metric.

The full trajectory-level loss is then defined as the sum of segment-level losses as:
\begin{equation}
\label{eqn:segmented_traj_loss}
\ell_{\text{traj}}(\tau^{\text{hyb}}, \tau^e) = \sum_{j=1}^n \ell_{\text{traj}}(\tau^{\text{hyb},j}, \tau^{e,j}).
\end{equation}

\subsection{Implementation Details}
\label{sec:method_details}

In this section, we detail the design of the fast control module $\pi^{\text{fast}}$ and the policy indicator module $I$. Our hybrid policy $\pi^{\text{hyb}}$ is designed to allocate tasks to the policies based on their complexity: $\pi^{\text{fast}}$ executes reach-and-place subtasks, while $\pi^{\text{vla}}$ handles active object manipulation. This division leverages the fact that contact-rich manipulation demands expressive, learned visuomotor control, whereas free-space transit is purely geometric and can be completed significantly faster using classical motion planning.

{\bf Fast control module $\pi^{\text{fast}}$.} The fast control module $\pi^{\text{fast}}$ is composed of two key components: a \emph{target pose predictor} $f_{\text{tgt}}$ and a \emph{low-level motion planner} $\mathcal{P}$.

The target pose predictor $f_{\text{tgt}}$ maps the current observation $\mathcal{O}_s$, language instruction $L$, and current proprioceptive state $\mathbf{p}_s$ to the target end-effector pose at the end of the current subtask (trajectory segment), denoted by $\mathbf{p}^{\text{tgt}}_s$ as:
\begin{equation}
\label{eqn:target_pose}
    \mathbf{p}^{\text{tgt}}_s = f_{\text{tgt}}(\mathcal{O}_s, L, \mathbf{p}_s).
\end{equation}

The low-level motion planner $\mathcal{P}$ then takes the current state $\mathbf{p}_s$, target pose $\mathbf{p}^{\text{tgt}}_s$, and current observation $\mathcal{O}_s$ to generate an action sequence $(a_s, \dots, a_{s+m})$ of horizon $m$ that moves the robot from $\mathbf{p}_s$ to $\mathbf{p}^{\text{tgt}}_s$ while avoiding collisions, i.e.,
\begin{equation}
\label{eqn:fast_traj}
    (a_s, \dots, a_{s+m}) = \mathcal{P}(\mathbf{p}_s, \mathbf{p}^{\text{tgt}}_s, \mathcal{O}_s).
\end{equation}

{\bf Target pose predictor $f_{\text{tgt}}$.} The target pose predictor identifies the relevant object for the current subtask and determines a collision-free hover pose directly above it. Specifically, from the observation $\mathcal{O}_s$, we first extract 3D bounding boxes $\mathcal{B} = \{\mathbf{b}_i\}_{i=1}^M$ for the objects in the scene, where $\mathbf{b}_i = (\mathbf{c}_i, \mathbf{e}_i)$ consists of 3D centroid $\mathbf{c}_i \in \mathbb{R}^3$ and 3D extents $\mathbf{e}_i \in \mathbb{R}^3$.

We score candidate object boxes by cross-attention between a learned positional embedding of each bounding box $\mathbf{b}_i$ (acting as the query) and the vision-language feature tokens extracted from the scene (acting as keys and values). More concretely, for the $i$-th object bounding box $\mathbf{b}_i$, its relevance score $r_i$ is computed as:
\begin{equation}
\label{eqn:box_score}
    r_i = \text{MLP}\left( \text{Softmax}\left(\frac{\mathbf{q}_i^{\text{box}} \mathbf{K}_{\text{vl}}^\top}{\sqrt{d_k}}\right) \mathbf{V}_{\text{vl}} \right),
\end{equation}
where $\mathbf{q}_i^{\text{box}} = \phi_{\text{pos}}(\mathbf{b}_i) \in \R^{d_k}$ is the learned positional embedding of the box, $\mathbf{K}_{\text{vl}}, \mathbf{V}_{\text{vl}} \in \R^{T \times d_k}$ are linear projections of the $T$ vision-language feature tokens $\mathbf{F}_{\text{vl}}$. A lightweight MLP head then maps the attended feature representation to a scalar score $r_i \in \mathbb{R}$. The target object is selected via $\hat{i} = \arg\max_i r_i$, and a top-down, collision-free hover pose above candidate box $\mathbf{b}_{\hat{i}}$ is assigned as the target pose $\mathbf{p}^{\text{tgt}}_s$.

We extract the vision-language features $\mathbf{F}_{\text{vl}}$ using the vision-language backbone of $\pi^{\text{vla}}$ to reuse its pretrained representations. Only the positional embedding $\phi_{\text{pos}}$, cross-attention projections, and the scoring MLP are trained from scratch. The target pose predictor is only called once at the beginning of every reach-and-place phase ($I=1$).

{\bf Low-level motion planner $\mathcal{P}$.} 
Our framework is agnostic to the planner choice, and we instantiate $\mathcal{P}$ using classical motion planners such as cuRoboV2 and VAMP.

{\bf Policy indicator module $I$.} Learning task boundaries directly from visual observations is an active research area. As in this work, we focus on hybridizing the VLA with a fast motion planner; we choose an effective and fast heuristic based on gripper state and target pose proximity to indicate the task boundaries. 

The policy begins in a reach-and-place phase ($I_0 = 1$), using $\pi^{\text{fast}}$ to navigate toward the target pose $\mathbf{p}^{\text{tgt}}_s$. Once the end-effector reaches the target ($d(\mathbf{p}_s^{ee}, \mathbf{p}^{\text{tgt}}_s) < \epsilon_d$, where $\epsilon_d > 0$ is some tolerance threshold), control is handed over to $\pi^{\text{vla}}$ ($I_s = 0$). 
Once active manipulation is complete, signaled by a toggle in the gripper state ($g_s \neq g_{s-1}$), the indicator switches back to $I_s = 1$, returning control to $\pi^{\text{fast}}$. Formally, let $g_s \in \{0, 1\}$ be the binary gripper state (open or closed) at step $s$, $\mathbf{p}_s^{ee}$ be the end-effector pose, and $\mathbf{p}^{\text{tgt}}_s$ be the active target pose. With initial state $I_0 = 1$, the indicator $I_s$ is recursively defined as:
\begin{equation}
\label{eqn:indicator}
I_s =
    \begin{cases}
        1, & \text{if } g_s \neq g_{s-1}, \\
        0, & \text{if } I_{s-1} = 1 \text{ and } d(\mathbf{p}_s^{ee}, \mathbf{p}^{\text{tgt}}_s) < \epsilon_d, \\
        I_{s-1}, & \text{otherwise.}
    \end{cases}
\end{equation}

\subsection{Training the Hybrid Architecture}
\label{sec:method_training}
Our hybrid policy design naturally simplifies the optimization in \equref{eqn:objective} and does not require end-to-end policy optimization. First, we freeze the pretrained $\pi^{\text{vla}}$, which already handles contact-rich manipulation. 
Furthermore, $\mathcal{P}$ guarantees reachability to any valid goal pose with minimal latency; the reach-and-place trajectory loss in \equref{eqn:segment_loss} is minimized whenever the target pose $\mathbf{p}^{\text{tgt}}_s$ is correctly predicted. Consequently, optimizing \equref{eqn:objective} reduces solely to training the target pose predictor $f_{\text{tgt}}$ (specifically its box-scoring head) to identify the target object.

{\bf Automated data generation and training.} To train the target scoring function without manual annotation, we extract supervision directly from expert demonstration trajectories. Reach-and-place segments correspond to free-space transit motions that precede an interaction. We detect the completion of each reach-and-place segment by identifying gripper events, namely any change in the gripper state across a threshold (closing to grasp or opening to release). At this gripper event, the target object or receptacle is identified by finding the bounding box closest to the robot's end-effector $\mathbf{p}^{ee}_{\text{event}}$ as: 
\be
i^* = \arg\min_i \|\mathbf{p}^{ee}_{\text{event}} - \mathbf{c}_i\|_2.
\ee
We then assign $i^*$ as the ground-truth target for all preceding states $\mathcal{O}_s$ within that reach-and-place segment. For any observation $\mathcal{O}_s$, given candidate bounding boxes $\mathcal{B}$ and frozen vision-language tokens $\mathbf{F}_{\text{vl}}$, the scoring head computes relevance scores $\{r_i\}_{i=1}^M$ according to \equref{eqn:box_score}. We optimize the trainable parameters of $f_{\text{tgt}}$ using a cross-entropy loss:
\begin{equation}
\label{eqn:tar_loss}
    \mathcal{L}_{\text{tgt}} = - \log \frac{\exp(r_{i^*})}{\sum_{j=1}^M \exp(r_j)}.
\end{equation}
This allows SkipVLA to learn robust target prediction across varying viewpoints throughout transit, requiring no additional data collection or manual annotations.

\definecolor{bestcell}{HTML}{DCE8F5}    %
\definecolor{secondcell}{HTML}{FDEAD6}  %
\providecommand{\rkfirst}[1]{\colorbox{bestcell}{#1}}
\providecommand{\rksecond}[1]{\colorbox{secondcell}{\underline{#1}}}

\begin{table}[t]
\centering
\setlength{\tabcolsep}{4pt}
\setlength{\fboxsep}{2pt}
\resizebox{\columnwidth}{!}{%
\begin{tabular}{lcccc}
\toprule
Policy & SR\,(\%)~$\uparrow$ & $n_{\text{succ}}$ & Wall Clock Time\,(s)~$\downarrow$ & VLA\,q~$\downarrow$ \\
\midrule
$\pi_{0.5}$ (base)           & $99.0$ & $495$ & $8.33{\pm}1.18$ ($8.3$) & $17.0{\pm}2.4$ ($17$) \\
\;+\,SkipVLA\,+\,VAMP         & $99.0$ & $495$ & \rksecond{$5.27{\pm}2.19$ ($4.4$)} & \rksecond{$10.6{\pm}4.7$ ($9$)} \\
\;+\,SkipVLA\,+\,cuRoboV2       & \rksecond{99.2} & \rksecond{496} & $6.28{\pm}2.82$ ($5.3$) & \rkfirst{$\mathbf{9.9{\pm}4.8}$ ($8$)} \\
\;+\,Heuristic Ablation       & \rkfirst{$\mathbf{99.6}$} & \rkfirst{$\mathbf{498}$} & \rkfirst{$\mathbf{5.11{\pm}1.16}$ ($4.7$)} & $10.7{\pm}2.2$ ($10$) \\
\midrule
SmolVLA (base)               & $39.0$ & $195$ & $12.61{\pm}4.99$ ($10.8$) & $21.3{\pm}8.3$ ($18$) \\
\;+\,SkipVLA\,+\,VAMP         & $77.0$ & $385$ & \rkfirst{$\mathbf{6.96{\pm}3.76}$ ($5.8$)} & \rksecond{$11.4{\pm}6.2$ ($10$)} \\
\;+\,SkipVLA\,+\,cuRoboV2       & \rkfirst{$\mathbf{82.0}$} & \rkfirst{$\mathbf{410}$} & $7.85{\pm}3.94$ ($6.8$) & \rkfirst{$\mathbf{10.8{\pm}6.5}$ ($9$)} \\
\;+\, Heuristic Ablation      & \rksecond{79.0} & \rksecond{395} & \rksecond{$7.25{\pm}3.02$ ($6.2$)} & $12.6{\pm}5.0$ ($11$) \\
\bottomrule
\end{tabular}}
\caption{\textbf{LIBERO-object Evaluation. }We evaluate SkipVLA across 10 pick and place tasks in simulation. We also ablate the indicator heuristics to show the model's adaptability. We calculate statistics over successful trials only. \textbf{Bold}: best; \underline{underline}: second-best in each backbone block. Numbers shown as Mean ${\pm}$ Standard Deviation (Median).}
\label{tab:libero_object_success}
\end{table}

\section{Experiments}
\label{sec:experiments}

\begin{figure*}[htbp]
\centering
\includegraphics[width=\textwidth]{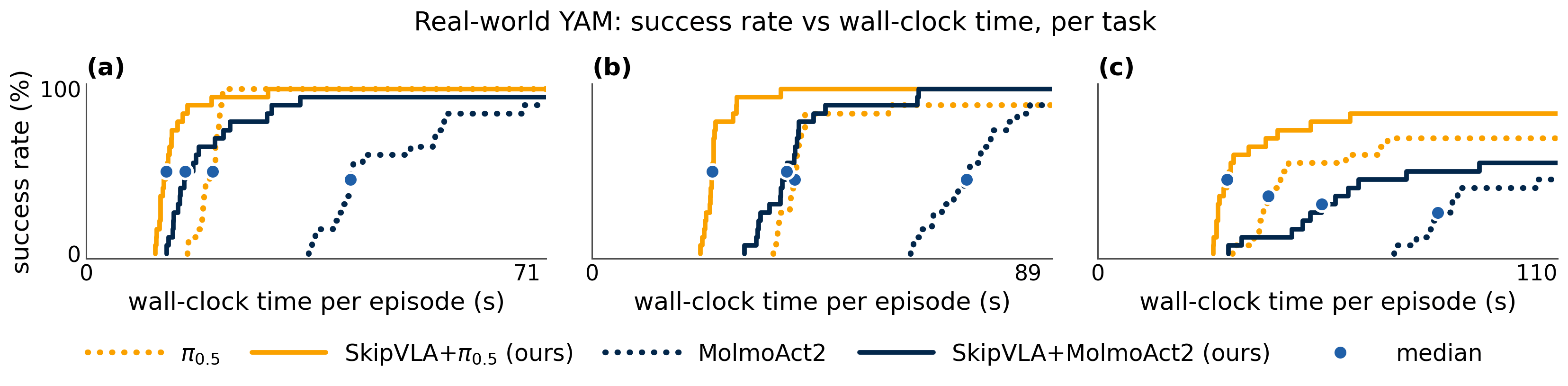}
\caption{Cumulative distribution of success rate (\%) vs. wall clock (s) for 3 tasks across $\pi_{0.5}$, MolmoAct2, $\pi_{0.5}$ + SkipVLA and MolmoAct2 + SkipVLA measured across 20 trials per task. \textbf{(a)} put the red block in the box. \textbf{(b)} put both the red and blue blocks in the box. \textbf{(c)} stack the 3 blocks in a tower.}
\label{fig:cdf}
\end{figure*}

Our evaluation investigates whether interleaving a fast, point-to-point classical motion planner with a VLA model achieves faster execution and better compute efficiency than purely end-to-end VLA execution without compromising on task success. Specifically, our experiments aim to answer two core questions:
\begin{itemize}
    \item \textbf{Efficiency:} Does offloading free-space transit to a classical planner significantly reduce wall-clock execution time and VLA query frequency across varying task horizons without compromising on task success?
    \item \textbf{Modularity:} Is the framework robust to variations in the underlying planner, latent context representation, and indicator function?
\end{itemize}

\subsection{Simulation Experiments}
\label{sec:sim_experiments}

On a standardized benchmark, we test for task speedup, VLA query reduction, and task completion.

{\bf Task setup \& benchmark.}
We evaluate on $13$ top-down pick-and-place tasks drawn from two suites in the LIBERO~\cite{liu2023libero} tabletop benchmark: \textsc{LIBERO-Object} (short-horizon, single-object manipulation) and \textsc{LIBERO-10} (longer-horizon multi-object manipulation). We deliberately select tasks solvable via top-down grasps since our objective is not to stress-test complex grasp synthesis, but to isolate and measure the efficiency gains of hybrid planning. We train a lightweight multi-head cross-attention scorer with a 3-layer MLP bounding box encoder on the human teleoperation demonstrations from LIBERO using all 50 demonstrations per task, 650 demonstrations in total.

\definecolor{bestcell}{HTML}{DCE8F5}    %
\definecolor{secondcell}{HTML}{FDEAD6}  %
\providecommand{\rkfirst}[1]{\colorbox{bestcell}{#1}}
\providecommand{\rksecond}[1]{\colorbox{secondcell}{\underline{#1}}}

\begin{table}[t]
\centering
\setlength{\tabcolsep}{3pt}
\setlength{\fboxsep}{2pt}
\resizebox{\columnwidth}{!}{%
\begin{tabular}{lcccc}
\toprule
Policy & SR\,(\%)~$\uparrow$ & $n_{\text{succ}}$ & Wall Clock Time\,(s)~$\downarrow$ & VLA\,q~$\downarrow$ \\
\midrule
$\pi_{0.5}$ (base)           & \rksecond{94.7} & \rksecond{142} & $15.69{\pm}2.35$ ($15.2$) & $31.7{\pm}4.7$ ($31$) \\
\;+\,SkipVLA\,+\,VAMP         & \rksecond{94.7} & \rksecond{142} & \rkfirst{$\mathbf{13.84{\pm}7.07}$ ($11.0$)} & \rksecond{$26.5{\pm}14.4$ ($21$)} \\
\;+\,SkipVLA\,+\,cuRoboV2       & \rkfirst{$\mathbf{95.3}$} & \rkfirst{$\mathbf{143}$} & $17.84{\pm}4.99$ ($16.3$) & \rkfirst{$\mathbf{24.1{\pm}9.9}$ ($21$)} \\
\;+\ Heuristic Ablation       & \rksecond{94.7} & \rksecond{142} & \rksecond{$15.28{\pm}6.02$ ($13.2$)} & $31.4{\pm}12.0$ ($27$) \\
\midrule
SmolVLA (base)               & $5.3$ & $8$ & $26.91{\pm}7.40$ ($23.7$) & $45.0{\pm}12.3$ ($39.5$) \\
\;+\,SkipVLA\,+\,VAMP         & $24.7$ & $37$ & \rkfirst{$\mathbf{15.29{\pm}9.17}$ ($11.3$)} & \rksecond{$24.6{\pm}15.8$ ($17$)} \\
\;+\,SkipVLA\,+\,cuRoboV2       & \rkfirst{$\mathbf{29.3}$} & \rkfirst{$\mathbf{44}$} & $17.97{\pm}8.02$ ($15.1$) & \rkfirst{$\mathbf{22.8{\pm}12.4}$ ($18.5$)} \\
\;+\ Heuristic Ablation      & \rksecond{26.7} & \rksecond{40} & \rksecond{$15.61{\pm}4.75$ ($14.2$)} & $26.6{\pm}7.8$ ($25$) \\
\bottomrule
\end{tabular}}
\caption{\textbf{LIBERO-10 Evaluation.} We evaluate SkipVLA across 3 pick and place tasks in simulation against 2 baselines. We also ablate the indicator heuristics to show the model's adaptability. We calculate statistics over successful trials only. \textbf{Bold}: best; \underline{underline}: second-best in each backbone block. Numbers shown as Mean ${\pm}$ Standard Deviation (Median).}
\label{tab:libero10_success}
\end{table}
 
{\bf Baselines \& evaluation metrics.}
As our base VLAs, we choose $\pi_{0.5}$~\cite{pi05}, a large-scale pretrained model with $3.5B$ parameters, and SmolVLA~\cite{smolvla25}, a comparatively smaller and faster model with $0.5B$ parameters.  For both of these VLAs, we use the publicly released fine-tuned weights on LIBERO. We experiment with using our SkipVLA framework on each VLA and report results using different planner algorithms, such as VAMP~\cite{vamp_2024} and cuRoboV2~\cite{curobo_v2}.

For evaluation, each method is repeated across $50$ trials per task, measuring success rate (SR, \%), total successful attempts ($n_{succ}$), wall-clock time (Wall s), and VLA query count (VLA q). We use a hard limit of 600 environment steps in MuJoCo as a timeout.

{\bf Results.}
We present the results in~\tabref{tab:libero_object_success} and~\tabref{tab:libero10_success}. First, we note that, regardless of the planner, SkipVLA matches the base model's performance. Across both VLA base models, SkipVLA with VAMP as the planner reduces the wall-clock time by $40\%$ and $27\%$ on LIBERO-object and LIBERO-10, respectively. We also see a proportional drop in the number of VLA queries, indicating that the time reduction is largely due to fewer VLA queries. We also note that SkipVLA shows consistent performance regardless of the planner choice. While cuRoboV2 has a slightly higher success rate, VAMP demonstrates a shorter wall-clock time.

\definecolor{bestcell}{HTML}{DCE8F5}
\newcommand{\best}[1]{\colorbox{bestcell}{$\mathbf{#1}$}}
\newcommand{\bestt}[1]{\colorbox{bestcell}{\textbf{#1}}}

\begin{table*}[t]
\centering
\setlength{\tabcolsep}{5pt}
\setlength{\fboxsep}{1.5pt}
\resizebox{\textwidth}{!}{%
\begin{tabular}{ll c cc cc cc}
\toprule
 & & & \multicolumn{2}{c}{\textbf{Wall Time (s)}~$\downarrow$} & \multicolumn{2}{c}{\textbf{VLA Calls}~$\downarrow$} & \multicolumn{2}{c}{\textbf{Energy (J)}~$\downarrow$} \\
\cmidrule(lr){4-5}\cmidrule(lr){6-7}\cmidrule(lr){8-9}
Task & Policy & SR~$\uparrow$ & mean\,$\pm$\,std & median & mean\,$\pm$\,std & median & mean\,$\pm$\,std & median \\
\midrule
\multirow{4}{*}{\shortstack[l]{\textbf{Single-Object}\\\textit{Put red block in box}}}
 & $\pi_{0.5}$ (base)          & \bestt{20/20} & $19.81 \pm 1.73$ & $20.33$ & $9.15 \pm 0.75$ & $9$ & $835.1 \pm 73.9$ & $845.3$ \\
 & \;+\,SkipVLA (ours)         & \bestt{20/20} & \best{14.11 \pm 4.16} & \best{12.91} & \best{6.10 \pm 1.55} & \best{6} & \best{577.9 \pm 167.1} & \best{539.0} \\
\cmidrule{2-9}
 & MolmoAct2 (base)            & $18/20$ & $46.56 \pm 9.53$ & $42.56$ & $18.06 \pm 3.44$ & $17$ & $1876.6 \pm 383.5$ & $1720.8$ \\
 & \;+\,SkipVLA (ours)         & \bestt{19/20} & \best{18.82 \pm 6.27} & \best{15.93} & \best{8.16 \pm 2.77} & \best{7} & \best{712.1 \pm 236.4} & \best{610.0} \\
\midrule
\multirow{4}{*}{\shortstack[l]{\textbf{Multi-Object}\\\textit{Put red \& blue in box}}}
 & $\pi_{0.5}$ (base)          & $18/20$ & $41.77 \pm 5.09$ & $41.36$ & $17.89 \pm 2.22$ & $18$ & $1727.0 \pm 214.5$ & $1710.0$ \\
 & \;+\,SkipVLA (ours)         & \bestt{20/20} & \best{25.53 \pm 3.72} & \best{24.56} & \best{11.15 \pm 1.53} & \best{11} & \best{1050.8 \pm 146.4} & \best{1017.0} \\
\cmidrule{2-9}
 & MolmoAct2 (base)            & $18/20$ & $76.04 \pm 7.17$ & $76.45$ & $29.00 \pm 2.72$ & $29$ & $3236.8 \pm 303.8$ & $3251.4$ \\
 & \;+\,SkipVLA (ours)         & \bestt{20/20} & \best{41.66 \pm 9.48} & \best{39.73} & \best{17.85 \pm 4.11} & \best{17} & \best{1562.2 \pm 355.3} & \best{1488.1} \\
\midrule
\multirow{4}{*}{\shortstack[l]{\textbf{Precision Stacking}\\\textit{Stack tower}}}
 & $\pi_{0.5}$ (base)          & $14/20$ & $47.62 \pm 11.91$ & $42.73$ & $20.79 \pm 5.40$ & $19$ & $1976.2 \pm 502.5$ & $1781.2$ \\
 & \;+\,SkipVLA (ours)         & \bestt{17/20} & \best{36.13 \pm 9.68} & \best{32.41} & \best{16.77 \pm 5.41} & \best{15} & \best{1631.4 \pm 476.0} & \best{1429.9} \\
\cmidrule{2-9}
 & MolmoAct2 (base)            & $9/20$ & $87.13 \pm 9.85$ & $85.26$ & $31.89 \pm 3.59$ & $31$ & $3829.6 \pm 792.3$ & $3621.2$ \\
 & \;+\,SkipVLA (ours)         & \bestt{11/20} & \best{58.11 \pm 17.75} & \best{56.19} & \best{25.82 \pm 8.12} & \best{25} & \best{2167.1 \pm 669.2} & \best{2112.4} \\
\bottomrule
\end{tabular}}
\caption{\textbf{Real-World Evaluation.} We evaluate SkipVLA against two baselines in the real world on a physical 6-DoF YAM robot arm across three tasks (20 trials per task). Statistics for wall-clock time, VLA calls, and energy are calculated only from successful trials. SR higher is better; all other metrics lower is better; \textbf{bold} and shading mark the better value within each backbone block.}
\label{tab:rw_yam_taskwise}
\vspace{-0.4cm}
\end{table*}

{\bf Ablation of indicator heuristics.}
We provide ablations on the heuristics to identify the task boundary. For this ablation, we use changes in object height as an indicator of the transition between reach-and-place and active manipulation; i.e., a change beyond a threshold denotes the transition. We choose Vamp as the planner. From~\tabref{tab:libero_object_success} and~\tabref{tab:libero10_success}, we observe that the height-based heuristic performs comparably to other configurations, showing the stability of SkipVLA. However, our proposed indicator heuristics with the VAMP planner yield the shortest overall wall clock time, indicating the fastest task completion.

\subsection{Real-World Experiments}
\label{sec:real_world}

{\bf Experiment setup.}
To validate our method in the real world, we deployed SkipVLA on a physical 6-DoF YAM manipulator. Our method is tested against two baselines across three tasks in the real world: $\pi_{0.5}$~\cite{pi05} and MolmoAct2~\cite{molmo26}, each fine-tuned on 50 teleoperated demonstrations per task.

We evaluate across three top-down pick-and-place tasks of increasing horizon complexity. In addition to success rate, wall-clock time, and VLA query count, we also measure the total energy consumption per trial on the NVIDIA Jetson Thor onboard compute platform. Each task is evaluated over 20 independent trials:
\begin{itemize}
\item \textbf{Single-Object:} Put the red block in the box (90-second timeout).
\item \textbf{Multi-Object:} Put the red and blue blocks in the box (120-second timeout).
\item \textbf{Multi-Object with precision placing:} Stack three blocks into a tower (120-second timeout).
\end{itemize}

{\bf Main results.}
Consistent with our simulation results, SkipVLA outperforms both $\pi_{0.5}$ and MolmoAct2 across all tasks and measured metrics, as shown in~\tabref{tab:rw_yam_taskwise}. 
Specifically, SkipVLA achieves substantial reductions in execution time across all three tasks: wall-clock completion time decreases by an average of $46.0\%$ on MolmoAct2 (up to $59.6\%$, a $2.47\times$ speedup) and $30.6\%$ on $\pi_{0.5}$. Concurrently, VLA queries drop by an average of $33.8\%$ across tasks (reducing calls by over half on MolmoAct2 for single-object), confirming that offloading free-space transit bypasses a large fraction of costly policy evaluations. Crucially, these efficiency gains come with matched or strictly improved success rates on every task: SkipVLA boosts multi-object placement from $90\%$ to a perfect $100\%$ on both backbones, and raises precision stacking success by $+15.0\%$ percentage points for $\pi_{0.5}$ and $+10.0\%$ percentage points for MolmoAct2.

Next, in~\figref{fig:cdf} we report our real-world results as a cumulative distribution that shows the fraction of episodes solved within a given wall-clock budget. As visible from the graphs, SkipVLA encloses a larger area under the curve than its baseline counterpart on every task, demonstrating that SkipVLA reaches a higher final success rate within a substantially smaller wall-clock budget. By bypassing the base VLA during free-space transit, the hybrid policy drastically reduces the number of required queries to the generative policies and offloads the work to a lightweight, fast, and efficient CPU-based motion planner (VAMP). 

{\bf Energy comparison.} As shown in~\tabref{tab:rw_yam_taskwise}, SkipVLA is also substantially more energy-efficient on the Jetson Thor compute module, cutting per-trial energy consumption by an average of $52.4\%$ on MolmoAct2  and $29.1\%$ on $\pi_{0.5}$. This confirms that our hybrid approach achieves higher physical throughput with a strictly lighter energy footprint.

{\bf Additional discussion.}
Although SkipVLA was designed primarily to reduce inference latency by minimizing VLA queries, we observe that it also improves task success rates, such as improving SmolVLA from $39.0\%$ to $82.0\%$ on \textsc{LIBERO-Object} and improving real-world precision stacking by approximately $20\%$. 
We attribute this gain to the planners, which limit compounding errors. 

Imitation learning bounds a policy's error as $\epsilon$ when its state remains within the experts' distribution. During policy rollout, an out-of-distribution shift in state compounds, incurring a cost bounded by $O(\epsilon T^2)$ over task horizon $T$~\cite{pmlr}. In an end-to-end VLA, a small error during free-space transit can displace the robot out of distribution before it reaches the object, causing grasp failures. SkipVLA mitigates this in two ways: first, restricting VLA queries to short active-manipulation segments of length $T_{\text{manip}} \ll T$ shrinks the time horizon over which errors compound. Second, by planning collision-free paths directly to a nominal hover pose above the object, the classical planner physically re-anchors the robot to in-distribution initial states before each active-manipulation subtask, preventing error propagation to subsequent subtasks.

Lastly, SkipVLA in this work is restricted to pick-and-place tasks due to design choices that were outside the scope of this project: the indicator $I$ depends on gripper-state transitions, which assumes that every active-manipulation segment ends in a grasp or release. Future work aims to make SkipVLA generalizable across all tasks by replacing the heuristic-based indicator with a learned, task-conditioned temporal segmentation module.

\section{Conclusion}
\label{sec:conclusion}

We present SkipVLA, a method for accelerating pretrained VLAs by offloading free-space motion to a classical motion planner.
SkipVLA uses the planner to move the robot to a predicted target pose, hands control to the VLA for contact-rich skills, and returns control to the planner when the gripper opens or closes.
A lightweight scoring head on the frozen vision-language backbone of the VLA selects the target object and is trained from existing demonstrations with labels extracted from gripper events.
Compared to $\pi_{0.5}$ and MolmoAct2 on a physical 6-DoF YAM arm, SkipVLA completes tasks up to 2.5x faster and uses up to less half the energy of the baseline policy, while matching or improving success rate on every task.
The speedup of SkipVLA is larger for the slower MolmoAct2 than for $\pi_{0.5}$ on every task.
In simulation, SkipVLA also raises the success rate of SmolVLA significantly on our LIBERO experiments, suggesting that shorter VLA segments limit compounding error.
The performance of SkipVLA demonstrates that VLAs can benefit from classical motion planning without retraining the policy.

\section*{Acknowledgment}
Gemini and Claude assisted with editing and grammar enhancement; all content was verified by the authors.
\printbibliography{}

\end{document}